\documentclass[letterpaper]{article} 
\usepackage{aaai25}  
\usepackage{times}  
\usepackage{helvet}  
\usepackage{courier}  
\usepackage[hyphens]{url}  
\usepackage{graphicx} 
\usepackage{natbib}  
\usepackage{caption} 
\usepackage{algorithm}
\usepackage{algorithmic}
\usepackage{booktabs}
\usepackage{amsmath}
\usepackage{multirow}
\usepackage{xcolor}

\usepackage{newfloat}
\usepackage{listings}
\DeclareCaptionStyle{ruled}{labelfont=normalfont,labelsep=colon,strut=off} 
\floatstyle{ruled}
\newfloat{listing}{tb}{lst}{}
\floatname{listing}{Listing}
\title{ALADE-SNN: Adaptive Logit Alignment in Dynamically Expandable Spiking Neural Networks for Class Incremental Learning}
\author {
    Wenyao Ni\textsuperscript{\rm 1,\rm 2},
    Jiangrong Shen\textsuperscript{\rm 1,\rm 2,\rm 3}
    Qi Xu\textsuperscript{\rm 4},
    Huajin Tang\textsuperscript{\rm 1,\rm 2}
}
\affiliations {
    \textsuperscript{\rm 1}State Key Lab of Brain-Machine Intelligence, Zhejiang University\\
    \textsuperscript{\rm 2}College of Computer Science and Technology, Zhejiang University\\
    \textsuperscript{\rm 3}School of Computer Science and Technology, Xi'an Jiaotong University\\
    \textsuperscript{\rm 4}School of Computer Science and Technology, Dalian University of Technology\\
}

\usepackage{bibentry}

\begin{document}

\maketitle

\begin{abstract}
Inspired by the human brain's ability to adapt to new tasks without erasing prior knowledge, we develop spiking neural networks (SNNs) with dynamic structures for Class Incremental Learning (CIL). Our comparative experiments reveal that limited datasets introduce biases in logits distributions among tasks. Fixed features from frozen past-task extractors can cause overfitting and hinder the learning of new tasks.
To address these challenges, we propose the ALADE-SNN framework, which includes adaptive logit alignment for balanced feature representation and OtoN suppression to manage weights mapping frozen old features to new classes during training, releasing them during fine-tuning. This approach dynamically adjusts the network architecture based on analytical observations, improving feature extraction and balancing performance between new and old tasks.
Experiment results show that ALADE-SNN achieves an average incremental accuracy of 75.42 ± 0.74\% on the CIFAR100-B0 benchmark over 10 incremental steps. ALADE-SNN not only matches the performance of DNN-based methods but also surpasses state-of-the-art SNN-based continual learning algorithms. This advancement enhances continual learning in neuromorphic computing, offering a brain-inspired, energy-efficient solution for real-time data processing.
\end{abstract}

\section{Introduction}
The human brain exemplifies a sophisticated system utilizing synaptic plasticity to adapt neural circuits to new tasks, without erasing previously stored knowledge during new learning \cite{bassett2011dynamic}. This ability, known as lifelong learning, is crucial for survival in dynamic environments \cite{yang2009stably, kudithipudi2022biological}. 
These processes involve the dynamic reorganization of neural circuits and the redistribution of neural activities, allowing the brain to allocate resources more efficiently in response to varying task complexities, thereby facilitating cognitive functions. However, current deep learning algorithms always cause the networks to quickly fall into catastrophic forgetting \cite{french1999catastrophic} when learning new tasks.
Inspired by the biological comuting processes, researchers aim to develop artificial neural networks that can mimic this ability. Spiking neural networks (SNNs), which more closely resemble the workings of biological neurons through their event-driven processing, efficient synaptic plasticity and rich spatiotemporal dynamics, present a promising avenue for achieving biologically plausible continual learning \cite{maass1997networks, pei2019towards, subbulakshmi2021biomimetic,shi2024spikingresformer,xu2024enhancing,yang2024spiking}. In this paper, we focus on designing novel continual learning method for SNNs, by leveraging insights from dynamic neural architectures and neural activity redistribution.




Among continual learning scenarios, class-incremental learning (CIL) requires the network agents to fit new tasks arriving incrementally and always to perform well among the whole encountered task space, which makes it the most challenging scenario. In CIL, the conflict between learning new knowledge and retaining old ones is crucial for the overall performance \cite{shen2024efficient}. Hence, it is necessary to optimize task-specific subspace when learning new classes without disturbing old classes.


To address the issue of catastrophic forgetting in CIL, several methods have been proposed for traditional deep neural networks. 
Wherein, dynamic network methods can achieve superior results in CIL scenarios by expanding network structures to adapt representation ability of new classes, thus eliminating dependencies on parameter masks and prior task information. 
Meanwhile, in the realm of SNNs, efforts to tackle continual learning challenges based on combinations of dynamic networks \cite{chen2022state,shi2024towards,shen2023esl,shen2021hybridsnn} and other data replay methods have led to innovative approaches, such as DSD-SNN and SOR-SNN \cite{han2023enhancing,han2023adaptive}.
Despite significant advancements, the problem of unbalanced task cognition, which is caused by the limited number of memory samples during the rehearsal process of the learned task, still hurts. Although some techniques have been used to alleviate that problem, those methods tend to introduce hyper-parameters related to tasks \cite{wang2022foster} or rely solely on rough intuition \cite{yan2021dynamically} that does not adequately address that problem and finally cause the insufficient effectiveness of those approaches. Addressing this imbalance is crucial for improving the efficacy of continual learning models.

\begin{figure*}[!t]
    \centering
    \includegraphics[width=0.8\textwidth]{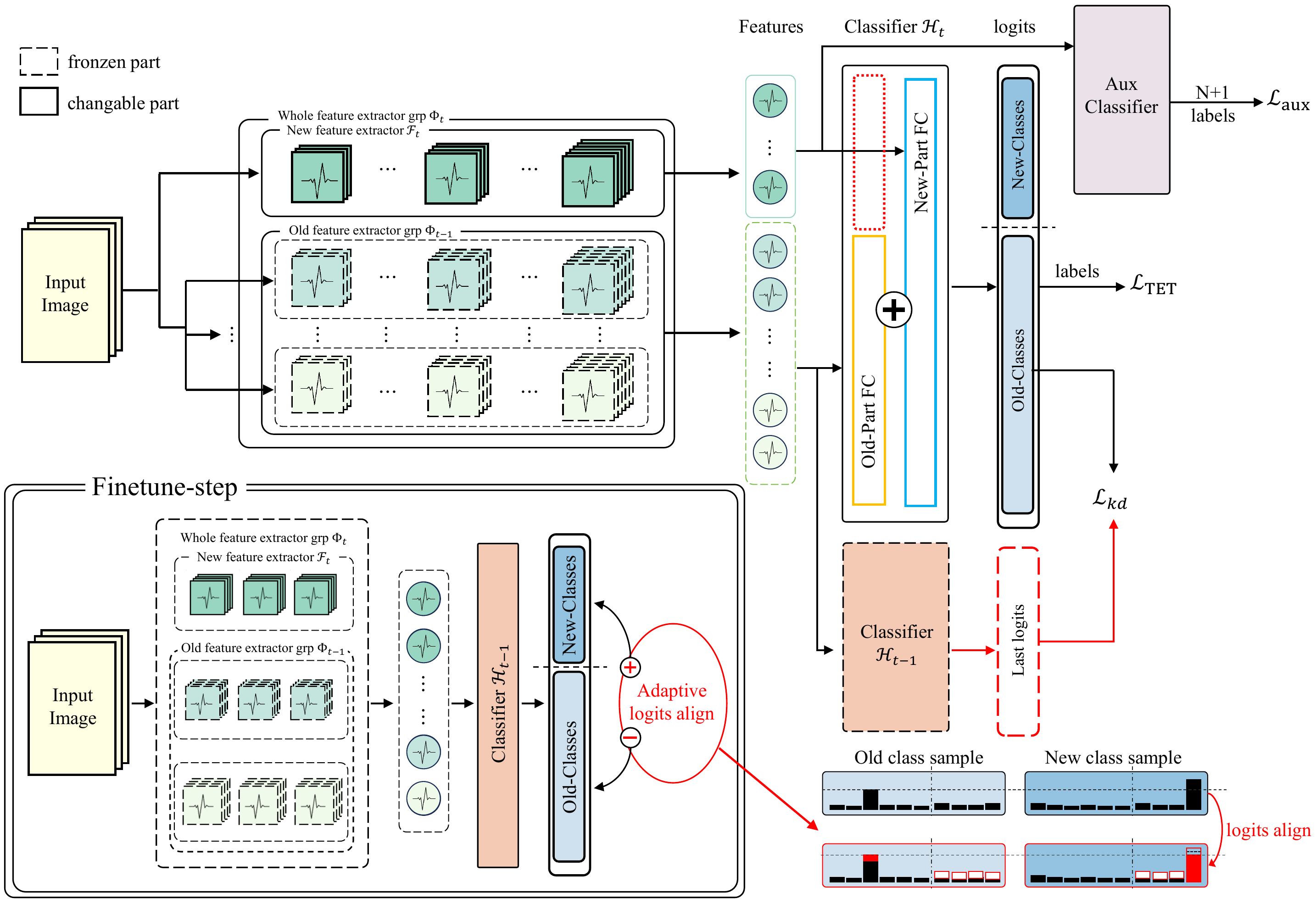}
    \caption{The overview of the proposed ALADE-SNN. This SNN-based continual learning framework includes two stages of training: the representation learning stage (upper part), where we implement OtoN suppression and knowledge distillation, and classifier learning stage (bottom left part), where we implement adaptive logit align to re-balance the knowledge distribution.}
    \label{fig:model_overview}
\end{figure*}

Therefore, based on the insights above, our research focuses on developing a novel CIL model tailored specifically for SNNs by leveraging insights from neural architecture reconfiguration and neural activity redistribution in the brain. 
As shown in Fig. \ref{fig:model_overview}, we have developed SNNs with dynamically expandable structures, designed to make a trade-off between stability and plasticity in CIL scenario. Additionally, we introduce an innovative adaptive logits alignment strategy aimed at mitigating the imbalance issues commonly encountered across varying tasks. This dual approach not only enhances the model's adaptability but also improves feature coherence among different tasks.
Our approach ensures that SNNs maintain long-term knowledge retention while efficiently incorporating new information, thereby advancing the state of continual learning in neuromorphic computing and real-time data processing applications.

The main contributions are summarized as follows:
\begin{itemize}
    \item Through comparative experiments, our study reveals that training with unbalanced datasets introduces additional bias in logit distributions
    . Fixed features from frozen extractors of past tasks cause incorrect dependencies, hindering the feature learning of new tasks.
    \item Consequently, we develop the dynamically expandable SNNs tailored for CIL, named ALADE-SNN. The framework incorporates adaptive logit alignment to balance the feature representation mapping among tasks. OtoN suppression is designed to suppress weights mapping frozen old features to new classes during training and release them during fine-tuning.
    \item The proposed ALADE-SNN achieves an average incremental accuracy of $75.42\%_{\pm0.74\%}$ on the CIFAR100-B0 with 10 steps. The experimental results demonstrate that the ALADE-SNN can attain comparable performance with DNN-based methods and surpass the state-of-the-art SNN-based continual learning algorithms.
\end{itemize}

\section{Related Works}
\textbf{Class-Incremental Learning with Architecture-based methods.} \ \ CIL aims to sequentially learn new classes while maintaining the ability to distinguish between classes from different learning episodes without relying on prior task information \cite{van2019three}. 
The architecture-based strategy expands network capacity dynamically to enhance the representation 
 capability for multiple tasks learning in CIL.
Some of these methods \cite{fernando2017pathnet, yoon2017lifelong, golkar2019continual, kang2022forget} require recording masks to select the right subsets during inference, which contradicts the setting in the CIL scenario. Consequently, DER \cite{yan2021dynamically} and simple-DER \cite{li2021preserving} eliminate the need for prior information and achieve continual learning by appending feature extractors along with tasks and fusing all features.
Several works \cite{douillard2022dytox, hu2023dense} have also proposed transformer-based methods that utilize the characteristics of transformers to implement continual learning.
Despite the progress in traditional neural networks, there is still a lack of sufficient exploration in applying these approaches to SNN-based models. The unique characteristics of spiking neural networks, such as their temporal dynamics and event-driven nature, present additional challenges and opportunities that have yet to be fully addressed in the context of continual learning.

\textbf{SNNs with Dynamic Structure for Continual Learning.} \ \ Our paper focuses on designing dynamic architectures that dynamically create or shift networks to fit specific tasks.  In SNN-related works, DSD-SNN \cite{han2023enhancing} proposes a method to dynamically adjust synapses between neurons in SNN to fit the continual tasks and SOR-SNN \cite{han2023adaptive} adopts a self-organize method that automatically chooses pathways according to the input information. Overall, the aforementioned methods do not place enough emphasis on the issue of data imbalance in each training stage in the CIL scenario.

\textbf{Classifier Learning with Class-imbalance Problem.} \ \ The class-imbalance problem during incremental learning has also been considered in relevant research. Some studies introduce additional training processes to solve that. For instance, BiC \cite{wu2019large} adds an extra correction layer trained with a separate validation set to correct the model's outputs. WA \cite{zhao2020maintaining} aligns the logits output according to the norms of the weight vectors of old and new classes. EEIL \cite{castro2018end} uses a balanced subset to fine-tune the classifier, and DER also adopts this technique.

Given the limitations of existing work, we focus on the imbalance problem in incremental learning and propose a more targeted method, ALADE-SNN, based on the results of comparative experiments to better equalize the cognitive levels of different categories. We deploy it on SNN and achieve comparable results with DNN-based and SNN-based methods.

\section{Analysis of the Feature Representation}

\begin{figure*}[t]
    \centering
    \includegraphics[width=0.8\textwidth]{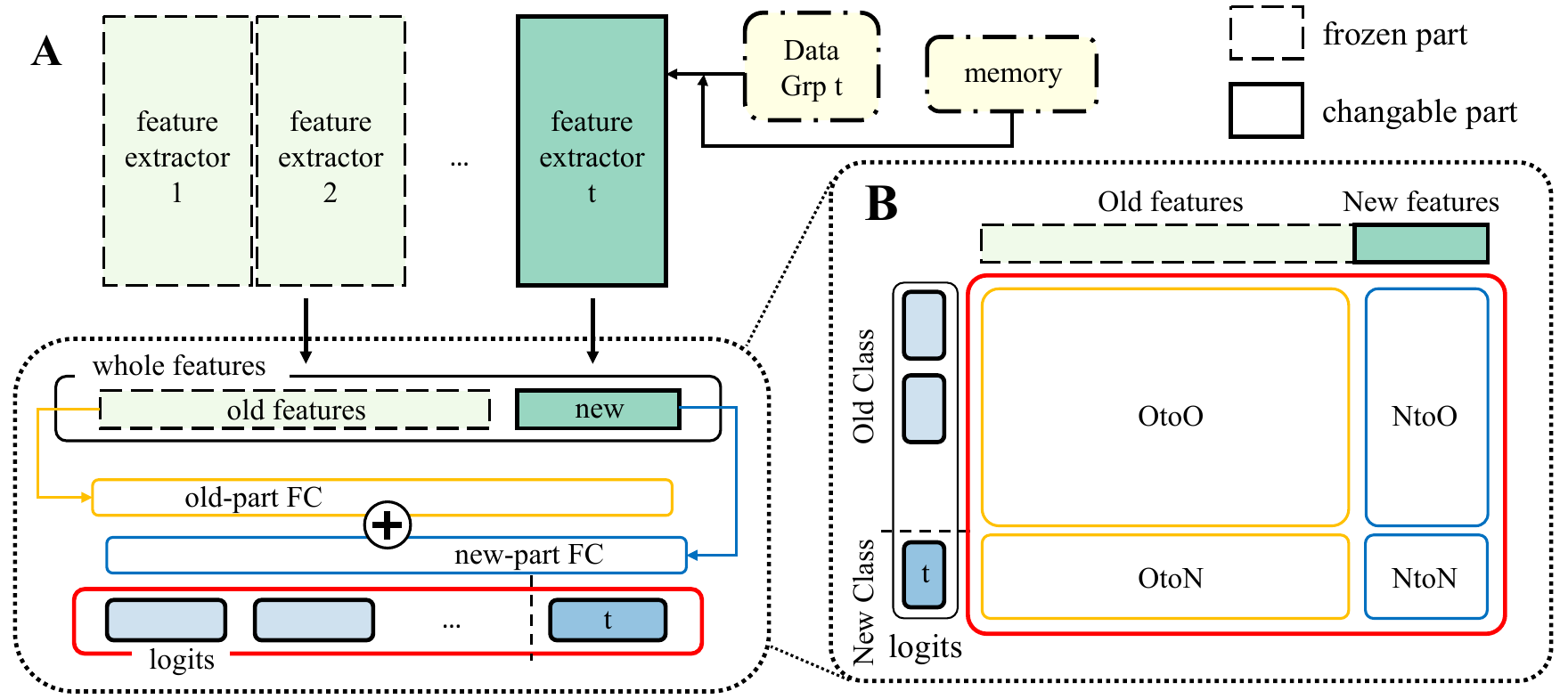}
    \caption{The architecture of the framework in comparative experiments. In these experiments, we primarily examine the impact of logits and weight on network performance. (A) The illustration of the overall network structure and its parameter properties. (B) The plot of the mapping relation between the separated deep features and the separated output logits.}
    \label{fig:toy_experiment_plotting}
\end{figure*}

This section analyzes the underlying reasons for catastrophic forgetting from the perspective of feature representation. Specifically, we conduct comparative experiments between an ideal incremental learner and typical incremental learning methods, such as DER, to better understand the potential structural characteristics of feature representation in an ideal incremental learner.
A more detailed description of the CIL tasks and experimental setting in the appendix can be found in the supplementary material.

In the experiments, we employ an Oracle model trained on all encountered data with single feature extractor as an optimal CIL learner, for its ability to continually learn different task samples and retain old knowledge without forgetting \cite{kim2023stability}. As depicted in Fig. \ref{fig:toy_experiment_plotting}, we compare the statistical characteristics of the Oracle model against other CIL models. These models are all built on DER's CIL framework but vary in terms of training data and weight limitations. The experiments are conducted on the CIFAR100 dataset, divided into 5 incremental learning steps, with a fixed memory size of 2000 samples, termed CIFAR100-B0-5steps. At each step, the models are evaluated based on their ability to retain knowledge from previous tasks while learning new ones. 

\begin{table*}[t]
    \caption{Statistical characteristics of the logits output of the comparative experiments. DER* here sets temperature=1 in fine-tuning step. Notably, "all-data" indicates that the model is trained on all observed data. When we suppress part of the connections in the final FC layer—i.e., zeroing the weights between different parts of features and predictions—we use the tag "no AtoB". The specific constraints for "AtoB" are shown in Fig. \ref{fig:toy_experiment_plotting}.}
    \label{tab:toy_exp_logit_plotting}
    \centering
    \begin{tabular}{l|cccc|cc}
    \toprule
        Exp & \textbf{1st avg/std} & \textbf{acc}(\%) & \textbf{2nd avg/std} & \textbf{acc}(\%) & \textbf{3st avg/std} & \textbf{acc} \\
    \midrule
        Oracle & 15.30 / 5.93 & 80.6 & 14.07 / 6.02 & 73.1 & 15.01 / 5.86 & 77.0 \\
    \midrule
        DER*(all-data) & 12.51 / 5.17 & 77.5 & 12.37 / 6.20 & 66.7 & \underline{14.69 / 6.87} & 76.8 \\
        DER*(all-data, no OtoN) & 10.90 / 4.92 & 77.4 & 10.79 / 6.07 & 69.0 & \underline{13.42 / 6.86} & 78.4 \\
        DER* & 10.45 / 4.60 & 68.3 & 10.45 / 5.12 & 67.0 & \underline{13.32 / 5.73} & 81.3 \\
    \bottomrule
    \end{tabular}
\end{table*}

\begin{table*}[t]
    \centering
    \caption{The performance of DER framework under varying training data settings and parameter constraints in the 4th step of the CIFAR100-B0-5steps. 
    "Fine-tune clf" refers to employing an additional step to fine-tune only the final classifier.
    }
    \label{tab:toy_exp_OtoN_plotting}
    
    \begin{tabular}{l|cc|l|cc}
    \toprule
        Exp (\textbf{no Fine-tune}) & \textbf{Avg}(\%) & \textbf{Last}(\%) & Exp (\textbf{Fine-tune clf}) & \textbf{Avg} & \textbf{Last}\\
    \midrule
        Oracle & 78.7 & 73.7 & \textbf{DER*(all-data, no OtoN)} & \textbf{79.7} & 75.5 \\
        DER*(all-data, no OtoN) & \textbf{77.1} & 70.9 & DER*(all-data, no NtoO) & 78.2 & 74.2  \\
        DER*(all-data) & 76.3 & 68.3 & DER*(all-data, no OtoN, NtoO) & 78.1 & 73.9 \\
    \midrule
        & & & DER* & 76.3 & 68.8 \\
    \bottomrule
    \end{tabular}
\end{table*}

Firstly, as shown in Table \ref{tab:toy_exp_logit_plotting}, we examine the distribution of logits values (model outputs before applying softmax) of the ground truth labels and the corresponding accuracy on the training dataset at the 3rd step of the CIFAR100-B0-5steps benchmark. This analysis helps us gain deeper insights into the bias induced by category imbalance \cite{castro2018end, zhao2020maintaining, wu2019large}. 
.
Results from the Oracle model show that when the training data from different tasks are plentiful and balanced, the network tends to produce a relatively consistent logits distribution among different tasks, as the difference in the average logits can be basically controlled within 1.3. The relatively low average logits of 2nd task may be attributed to fluctuations in the complexity and difficulty of different tasks.
However, under the DER framework, the logits outputs of old tasks tend to be weaker than those of the newly learned tasks. This occurs even though DER employs a fine-tuning step to address the imbalance problem, and DER (all-data) can access all previously seen data.

Furthermore, as shown in Table \ref{tab:toy_exp_OtoN_plotting}, we compute the average incremental accuracy at the 4th step of the same benchmark and observe a more peculiar phenomenon. Despite using the same balanced training set, the performance of DER (all-data) is worse than that of the Oracle (with 2.4\% decay in Avg and 5.4\% decay in Last), even though DER (all-data) retains more parameters due to several trained, frozen feature extractors. When we suppress the OtoN weights, this phenomenon is slightly alleviated as both Avg and Last have improved. If we further take an extra step to freeze all feature extractors and fine-tune the classifier at the end of each step (the bold term at the second column), models with more parameters ultimately perform better than the Oracle setting.

Through the aforementioned comparative experiments, we make the following speculations:

\begin{itemize}
    \item When trained with a sufficient and balanced dataset, the distribution of logits of different classes tends to be relatively balanced and may fluctuate depending on the difficulty of the tasks. Conversely, an imbalanced dataset can introduce additional bias in the logits values.
    \item The fixed features from the frozen feature extractors trained on prior tasks can introduce incorrect dependencies during the training process, potentially leading to overfitting and hindering the network from fully learning the features of new tasks.
\end{itemize}

Obviously, a simple fine-tune step under down-sampled, balanced dataset is not enough to fill the gap induced by imbalanced dataset. So based on the speculations above, we make targeted adjustments and design the adaptive logits alignment for SNNs.

\section{Dynamically Expandable SNNs with Adaptive Logits Alignment}
In this section, we introduce several components of ALADE-SNN according to the inferences from the aforementioned comparative experiments and explain in detail how these components function. Basically, our method follows DER framework, that is, to append new feature extractors and freeze the trained ones along the new task's coming, and make improvements.

\textbf{Adaptive Logits Alignment. }
Prior related works, such as \cite{yan2021dynamically, wang2022foster}, have proposed methods to solve the problem of bias in model outputs. For example, adding a scale factor to different parts of the logits to augment the memory of the old samples \cite{wang2022foster} or using a balanced fine-tuning method \cite{yan2021dynamically}. However, the former methods require extra hyper-parameters that tend to vary across different datasets, and the latter methods are not sufficiently effective, as shown in the comparative experiments above.


Based on the speculation above, we can further mitigate cognition bias by balancing the average ground-truth logits value of different tasks. We achieve this by manually adding a correction term to increase the logits of the new tasks during an extra step to fine-tune only the classifier under a balanced subset. At the incremental step $s$, given an image $x \in \hat{\mathcal{D}}_s$ and current feature extractor $\Phi_s$, classifier $\mathcal{H}_s$, the aligned logits values will be calculated as follows:

\begin{equation}
    \label{logits_calculation}
    O = \mathcal{H}_s(\Phi_s(x)) + \gamma * \text{mask}_s
\end{equation}

$\gamma$ is the correction term we add, and $\text{mask}_s$ is a mask of shape (classes, ) where the value for old classes is 0 and for other classes is 1. $O$ is the logits value after correction, which will participate in the prediction and loss calculation. 
In this way, due to the increased prediction possibility of ground-true label, the model loss of new task's training samples will be lower than the actual loss they can generate. Hence, the gradient balance in the original model will be disrupted and automatically lead the network to increase the logit of old tasks or decrease the new task's.

Regarding the setting of the value for $\gamma$, we adopt an adaptive approach to gradually approximate the logits of the old and new tasks to reach a balance shown in the comparative experiments. The pseudocode is shown in Alg. \ref{alg:logits_align}.

\begin{algorithm}[!h]
    \caption{Correction term $\gamma_e$ calculation each epoch}
    \label{alg:logits_align}
    \renewcommand{\algorithmicrequire}{\textbf{Input:}}
    \renewcommand{\algorithmicensure}{\textbf{Output:}}
    \begin{algorithmic}[1]
        \REQUIRE $\gamma_{e-1}$, $\Delta_{e-1}$, hyper-param $\alpha$, $\beta$, 
        Balanced data $\hat{\mathcal{D}}^{balance}_s$ downsampled from $\hat{\mathcal{D}}_s$
        \ENSURE The correction term $\gamma_{e}$, The logits difference $\Delta_{e}$ at this epoch

        \FOR{$batch$ in $\hat{\mathcal{D}}^{balance}_s$}
            \STATE Calculate logits using $\gamma_{e-1}$ as Ep. \ref{logits_calculation}
            \STATE Calculate loss and back-propogate it.
            \STATE $s_{new} \gets s_{new} + logits[y \in \mathcal{Y}_s, y]$
            \STATE $s_{old} \gets s_{old} + logits[y \in \mathcal{Y}_{1:s-1}, y]$
        \ENDFOR
        \STATE $\Delta_{e} \gets s_{new} / n_{new}$ - $s_{old} / n_{old}$
        \IF{$\Delta_{e} \cdot \Delta_{e-1} < 0$  or  $ (|\Delta_{st} - \Delta_{e}| > |\Delta_{st}| / \alpha$ and $ |\Delta_{e-1} -\Delta_{e}| < \delta_{\Delta} / \beta) $ }
            \STATE Fix correction term: $\gamma_{e} \gets \gamma_{e-1} + \Delta_{e}$
            \STATE Reset difference variation and start difference: $\delta_{\Delta} \gets 0$; $ \Delta_{st} \gets \Delta_{e} $
        \ELSE
            \STATE Record difference variation: $\delta_{\Delta} \gets Max(|\Delta_{e-1}-\Delta_{e}|, \delta_{\Delta})$
        \ENDIF
    \end{algorithmic}
\end{algorithm}

Since the added correction term usually does not have an immediate effect, we introduce two hyper-parameters, $\alpha$ and $\beta$, in Alg. \ref{alg:logits_align} to prevent correction term from oscillating caused by frequent modifications of $\gamma$. Specifically, $\alpha$ is used to indicate when the correction has already shown an impact, while $\beta$ is used to indicate when the effect of the correction term has become relatively weak. Though relatively slow and rough, this component does steadily reduce the bias between different tasks.

Despite the introduction of new hyper-parameters, they have little to do with loss calculation or the dataset itself. Therefore, we casually set $\alpha=8$ and $\beta=4$, which is sufficient to meet the experimental requirements. 

We apply this correction term only on an extra classifier learning stage, where all the feature extractors have been frozen, rather than directly applied during the train of the appended feature extractor $\mathcal{F}_s$. It is due to the relatively insufficient benefit when using this component to align logits during training of the feature extractor $\mathcal{F}_s$ in testing experiment. 
This may be because the proportion of exemplars is very small when training the feature extractor $\mathcal{F}_s$, and the correction term on logits actually leads to the insufficient feature extraction of $\mathcal{F}_s$, or the influence of the frozen representation mentioned in the comparative experiment.

Additionally, this component does not introduce new network structures or losses, so it can be used in parallel with other CIL methods/frameworks.

\textbf{OtoN Suppression. }
There exists a context prior that may introduce bias when transferred to the next task in the training data \cite{deng2021comprehensive}. The results of comparative experiments have also shown that dependencies formed by past features during learning can impact the network's generalization ability. 

To block the interference of frozen representation on new task's fitting, as been verified before, we manually zero the weight of the classifier mapping the frozen old features to new classes during the new feature extractor's training process. During the extra fine-tuning of classifier, this weight-constraint will be released to better utilize the established feature.

\textbf{Knowledge Distillation. }
Some related works utilized knowledge distillation \cite{hinton2015distilling} losses to provide more supervision for old classes \cite{wang2022foster, xu2024reversing,xu2023constructing}. In our experiments, we also observed that neural networks tend to reproduce the confidence distribution of old samples on old classes when undergoing incremental learning with sufficient data. Thus, we introduce distillation loss $\mathcal{L}_{kd}$ in our method. With the frozen feature extractors, theoretically, we only need previous classifier to obtain the confidence level at previous stage.

\begin{table*}[t]
    \centering
    \caption{Comparison of CIL approaches on CIFAR100-B0 benchmarks (averaged over three runs). \textit{Avg} denotes the average incremental accuracy (\%) over steps, and \textit{Last} represents the accuracy of the final step in each benchmark.}
    \label{cil_b0_table}

    \begin{tabular}{l|cc|cc|cc}
    \toprule
        \multirow{2}{*}{\textbf{Methods}} &\multicolumn{2}{c}{\textbf{CIFAR100-5steps}} & \multicolumn{2}{c}{\quad \textbf{CIFAR100-10steps} \quad } & \multicolumn{2}{c}{\quad \textbf{CIFAR100-20steps} \quad} \\
        \cline{2-7}
        & \textbf{Avg}(\%) & \textbf{Last}(\%) & \textbf{Avg}(\%) & \textbf{Last}(\%) & \textbf{Avg}(\%) & \textbf{Last}(\%) \\
    \midrule
        ResNet18 bound & - & 80.40 & - & 80.41 & - & 81.49 \\
    \midrule
        iCaRL\cite{rebuffi2017icarl} & 71.14 & - & 65.27 & 50.74 & 61.20 & 43.75  \\
        UCIR\cite{hou2019learning} & 62.77 & - & 58.66 & 43.39 & 58.17 & 40.63 \\
        BiC\cite{hou2019learning} & 73.10 & - & 68.80 & 53.54 & 66.48 & 47.02 \\
        WA\cite{zhao2020maintaining} & 72.81 & - & 69.46 & 53.78 & 67.33 & 47.31 \\
        PODNet\cite{douillard2020podnet} & 66.70 & - & 58.03 & 41.05 & 53.97 & 35.02 \\
        RPSNet\cite{rajasegaran2019random} & 70.50 & - & 68.60 & 57.05 & - & \\
        DER\cite{yan2021dynamically} & 76.80 & - & 75.36 & 65.22 & 74.09 & 62.48 \\
        Dytox+\cite{douillard2022dytox} & - & - & 75.54 & 62.06 & \textbf{75.04} & 60.03 \\
        TCIL\cite{huang2023resolving} & 77.70 & - & \textbf{77.30} & - & - & - \\
        DNE-1head\cite{hu2023dense} & - & - & 74.03 & \textbf{69.10} & 73.27 & \textbf{67.61} \\
    \midrule
        DSD-SNN\cite{han2023enhancing} & - & - & 60.47 & - & 57.39 & - \\
        Ours(ALADE-SNN) & \textbf{78.67} & 69.67 & \textbf{75.42} & \textbf{63.13} & \textbf{72.73} & \textbf{59.43} \\
    \bottomrule
    \end{tabular}
\end{table*}

\begin{table*}[t]
    \caption{Comparison of CIL approaches on CIFAR100-B50 benchmarks (averaged over three runs). }
    \label{cil_b50_tab}

    \centering
    \begin{tabular}{l|cc|cc}
    \toprule
        \multirow{2}{*}{\textbf{Methods}} & \multicolumn{2}{c}{ \textbf{CIFAR100-B50-5steps} } & \multicolumn{2}{c}{ \textbf{CIFAR100-B50-10steps} } \\
        \cline{2-5}
        & \textbf{Avg}(\%) & \textbf{Last}(\%) & \textbf{Avg}(\%) & \textbf{Last}(\%) \\
    \midrule
        ResNet18 bound & - & 79.89 & - & 79.91 \\
    \midrule
        iCaRL\cite{rebuffi2017icarl} & 65.06 & 56.07 & 58.59 & 49.43  \\
        UCIR\cite{hou2019learning} & 64.28 & 51.87 & 59.92 & 48.11 \\
        BiC\cite{hou2019learning} & 66.62 & 55.12 & 60.25 & 48.72 \\
        WA\cite{zhao2020maintaining} & 64.01 & 52.74 & 57.86 & 48.06 \\
        PODNet\cite{douillard2020podnet} & 67.25 & 55.94 & 64.04 & 51.65 \\
        DER\cite{yan2021dynamically} & 73.21 & 66.25 & \textbf{72.81} & \textbf{65.61} \\
    \midrule
        Ours(ALADE-SNN) & \textbf{74.50} & \textbf{66.9} & 71.50 & 63.3 \\
    \bottomrule
    \end{tabular}
\end{table*}

\textbf{SNN Setting. }
To enhance the implementation of image-classification tasks in SNN \cite{wu2021tandem,wang2023spatial,zhang2021rectified}, we introduce the SNN basic training method \cite{fang2021deep,fang2023spikingjelly} and the TET loss $\mathcal{L}_{TET}$ proposed in \cite{deng2021temporal}, which calculates cross-entropy loss and an extra regularization term, which is specifically calculated by MSE to reduce the risk of outliers in SNN. At every time step, the loss for SNNs is:

\begin{equation}
    \label{TET_loss_function}
    \mathcal{L}_{TET} = \frac{1}{T} \cdot \sum^T_{t=1}\mathcal{L}_{CE}[O(t), y] + \lambda \cdot \frac{1}{T} \sum^T_{t=1}MSE(O(t), \phi) 
\end{equation}

It should be noted that, since the correction term $\gamma$ is only used to re-align the logits, we only apply $\gamma$ in logits when calculating classification loss during classifier fine-tuning, not for possibly regularization or other loss.

Besides, we append the auxiliary loss $\mathcal{L}_{aux\_TET}$ proposed in the DER \cite{yan2021dynamically}. This loss leverages features from the current extractor $\mathcal{F}_{s}$ to predict the classes of the current task along with an additional class representing all previous classes (a total of $n + 1$ classes). We substitute the cross-entropy loss with the TET loss. The final loss function of our method during the training of $\mathcal{F}_s$ is presented as:

\begin{equation}
    \label{loss_function}
    \mathcal{L}_{ALADE} = \mathcal{L}_{TET} + \mathcal{L}_{aux\_TET} + \mathcal{L}_{kd}
\end{equation}

In classifier learning stage, since the whole group of feature extractors $\Phi_{s}$ has been frozen, we only utilize rate-based cross-entropy loss to redistribute the neural activity.

\begin{figure*}[h]
    \centering
    \includegraphics[width=0.9\textwidth]{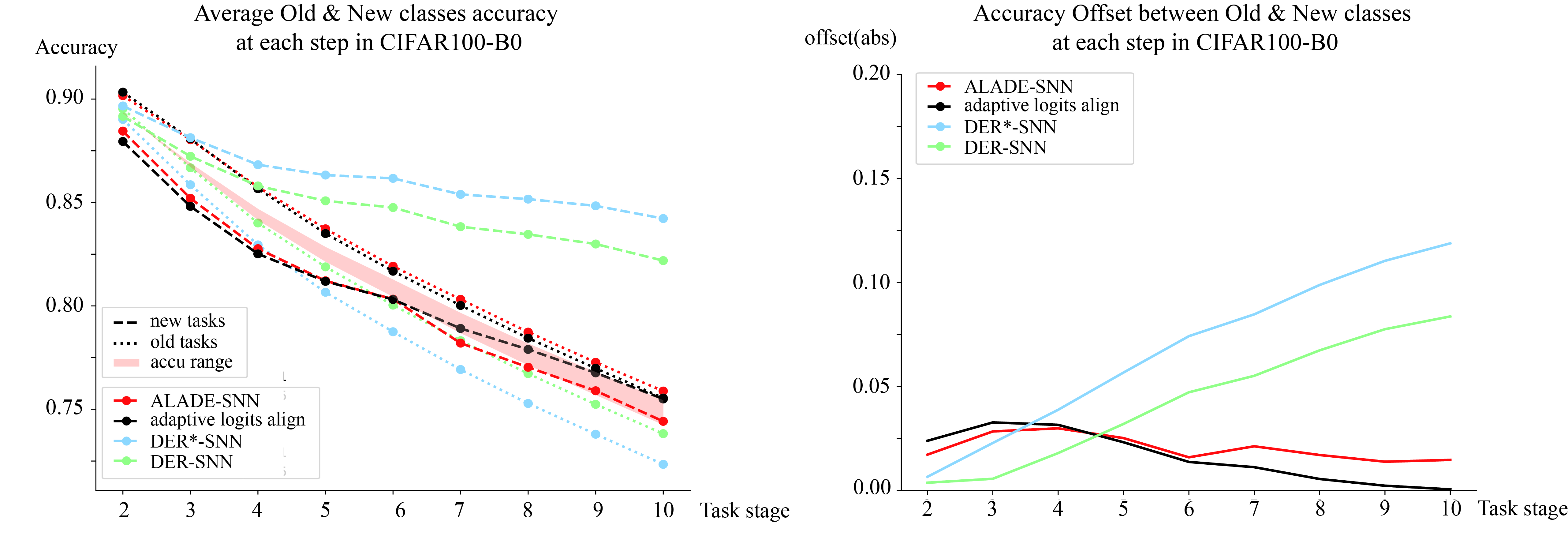}
    \caption{(Left) The plot of the curve of average accuracy (averaged across all steps) of old classes and new classes under different experimental settings in CIFAR100-B0 with 10 steps. (Right) The corresponding curve of absolute difference.}
    \label{fig:ablation_subplotting}
\end{figure*}

\section{Results}
To validate the effectiveness of our method, we conduct experiments primarily on CIFAR100 \cite{rebuffi2017icarl} using several benchmark protocols. Additionally, we perform a series of ablation experiments to evaluate the importance of different components. 
\textbf{Comparison of Performance. }
Table \ref{cil_b0_table} presents the results across CIFAR100-B0 benchmarks. On the one hand, ALADE-SNN outperforms the recently SNN-related work DSD-SNN, improving the average incremental accuracy from 60.47\% to 75.42\% (+14.95\%) under the incremental setting of 10 steps. Additionally, it enhances the incremental accuracy by 15.44\% under the setting of 20 steps, showcasing the superiority of our method in mitigating forgetting when applied to SNNs. When compared with ANN-based methods, ALADE-SNN still achieves comparable performance. Specifically, under the setting of 10 steps, our method achieves significantly better average incremental results compared to most ANN-based methods, lagging slightly behind the attention-based methods (Dytox and TCIL). However, in terms of the final accuracy, ALADE-SNN exhibits relatively poor performance, especially as the number of split steps increases. Under the setting of 20 steps, though still better than the other SNN-based method, ALADE-SNN shows a decline and still has a certain distance from recent ANN-based methods. This could be attributed to the SNNs' relatively weaker fitting and generalization abilities compared to ANN networks, given the small time window we have set. Consequently, this performance gap is likely to widen as the incremental learning process advances. This disparity explains why the performance of SNNs degrades more rapidly with an increase in split steps compared to ANNs. Anyway, under the setting of 5 steps, a superior improvement can be seen compared to the record of ANN-based methods ($\sim$1.80\% better than DER).

Similar trends are observed in the results of the CIFAR100-B50 benchmarks, as shown in Table \ref{cil_b50_tab}. Under the setting of 5 steps, both the average incremental accuracy and the last accuracy of our methods surpass those of the compared ANN-based methods ($\sim$1.30\% in average incremental accuracy and $\sim$0.7\% in last accuracy). And under the setting of 10 steps, just as revealed in the CIFAR100-B0 benchmarks, ALADE-SNN's performance falls behind that of typical ANN-based method DER. Though the narrow time window affects the fitting ability of SNN, it enables the network to achieve lower 
theoretical energy consumption when combining with neuromorphic hardware \cite{pei2019towards, imam2020rapid}(With estimation \cite{qiao2015reconfigurable}, 1.7625$\mu$J vs 3.367pJ under the same architecture as Resnet18 on CIFAR100 with T=4 for SNN as a reference). 

In addition, we also conduct experiments on neuromorphic dataset of DVS-CIFAR10. we set time window to be 10 in SNN according to \cite{deng2021temporal} and the memory size to be 1000. The whole dataset is only split to be 2 tasks (with 5 categories per task). ALADE-SNN achieves 83.5\%(Avg) and 77.2\%(Last) under Spiking-Resnet18, which is better than 81.4\%(Avg) and 74.8\%(Last) of DER under Resnet18. 

\textbf{Ablation experiment. }
To demonstrate how our method better balances the network's learning of old and new tasks and to validate the effectiveness of its components, we conducted several ablation experiments on CIFAR100-B0 with 10 steps. Table \ref{cil_ablation_experiment} summarizes the performance of various combinations of different components in our method. When comparing the direct application of the DER method in SNN with our approach, we observe an increase in the average incremental accuracy from 74.22\% to 75.42\%, indicating that our components can further enhance the overall performance in incremental learning. Notably, the adaptive logits alignment contributes significantly to this improvement, accounting for almost 1.0\% of the gain. 
Moreover, both OtoN suppression and knowledge distillation have played a certain role in improving the performance. Though OtoN suppression does not have the expected effect in the preliminary experiment, it still brings an improvement of $\sim$0.4\%. This discrepancy could be attributed to the limited sample scale of the old task in the CIL scenario, which may not provide as much misleading information as shown in the analytical experiment. Knowledge distillation also brings a slight gain when applied directly, though there is no significant gain when used with other components. The impact of knowledge distillation may be more pronounced when dealing with larger datasets in a single step, yet it continues to enhance the overall stability of performance.

\begin{table}
    \centering
    \caption{The ablation experiments of ALADE-SNN of the CIFAR100-B0-10steps (under three seeds' run)}
    \vspace{4pt}
    \label{cil_ablation_experiment}
    \begin{tabular}{l|c|c}
        \toprule
            \textbf{Components setting} & \textbf{Avg}(\%) & \textbf{Last}(\%) \\
        \midrule
            ALADE-SNN & $75.42_{\pm 0.74}$ & $63.13_{\pm 0.4}$ \\
        \midrule
            - OtoN suppression & $75.30_{\pm 1.20}$  & $62.30_{\pm 0.94}$ \\
            - knowledge distillation & $75.51_{\pm 1.31}$ & $63.60_{\pm 0.75}$ \\
            - adaptive logit alignment & $74.51_{\pm 0.76}$ & $61.80_{\pm 0.78}$ \\
        \midrule
            + adaptive logit alignment & $75.25_{\pm 1.09}$ & $62.77_{\pm 0.55}$ \\
            + OtoN suppression & $74.65_{\pm 1.44}$ & $62.03_{\pm 0.40}$ \\
            + knowledge distillation & $74.47_{\pm 1.85}$ & $61.70_{\pm 1.60}$ \\
            DER-SNN* & $74.22_{\pm1.06}$ & $61.20_{\pm0.56}$ \\
        \bottomrule
    \end{tabular}
\end{table}

To provide a clearer demonstration of how our method enhances overall performance by balancing the classes of different tasks in the network, we visualized the average accuracy changes for new and old tasks in the ablation experiments. As depicted in Fig. \ref{fig:ablation_subplotting}, when utilizing adaptive logit alignment, the average accuracy for both new and old tasks tends to converge around the overall incremental accuracy, while the performance gap between old and new tasks gradually widens with the addition of incremental tasks. This divergence is more distinctly illustrated by the curves of absolute difference in Fig. \ref{fig:ablation_subplotting} (right).
It should be noted that the "all" setting does not seem to perform as effectively as adaptive logit alignment in reducing such discrepancies. This result could be attributed to other components themselves having the capability to enhance memory recall, potentially leading to a slightly excessive performance improvement of old tasks. Compared to the difference in accuracy of DER between old and new tasks, this interval is acceptable.

\section{Conclusion}
Overall, we proposed ALADE-SNNs to enhance continual learning by adapting the dynamic network structure and rebalancing the cognitive load of old and new tasks. Through a series of experiments, we have demonstrated its efficacy in continual learning within CIL scenarios and validated the effectiveness of all components. Consequently, this method expands the utilization of SNNs in mitigating catastrophic forgetting within CIL contexts. 

\section{Acknowledgments}
This work was supported by National Natural
Science Foundation of China under Grant (No.62236007, No.62306274, No.32441113, No. 62206037, No. 62476035, No.61925603), Key R$\&$D Program of Zhejiang (2023C03001), and the Open Research Fund of the State Key Laboratory of Brain-Machine Intelligence, Zhejiang University (Grant No. BMI2400012).


\newpage
\appendix

\section{Supplemenary Related works}

Current methods for CIL mainly fall into three categories: rehearsal-based, Constraints-based  and architecture-based methods \cite{yang2023cross}. Except for the architecture-based strategy as mentioned above, the other two categories are as follows:

\textbf{Rehearsal-based methods.} The rehearsal technique commonly preserves a limited portion of data from prior tasks or replays previous examples using generative models while training new classes. 
Within the realm of rehearsal-based methods, certain approaches employ compression techniques to store more samples within a restricted memory capacity \cite{iscen2020memory, hayes2020remind, luo2023class}. Some methods rely on a single prototype sample for each class \cite{petit2023fetril, zhu2022self}. 
The majority of the aforementioned techniques necessitate the preservation of samples from previous instances, which could potentially infringe upon user privacy in some situations.
Alternative methods employ additional models to produce pseudo-samples, hence strengthening the retention of previously acquired tasks \cite{shin2017continual, lesort2019generative, van2020brain}.
However, the process of training generative models for the pseudo-samples brings extra computation and storage burden for these techniques.


\textbf{Constraints-based methods. }
Additionally, several methods impose constraints on the model to retain the necessary components for old tasks. These constraints can be applied to intermediate features \cite{douillard2020podnet}, prediction probabilities \cite{rebuffi2017icarl, castro2018end}, and model weights \cite{kirkpatrick2017overcoming, zenke2017continual, kang2022class}. 
Elastic Weight Consolidation (EWC) and Synaptic Intelligence (SI) impose penalties on the model weights to preserve the importance of the weights associated with previous tasks while training new tasks \cite{kirkpatrick2017overcoming, zenke2017continual}.
Knowledge distillation is a widely employed technique that implements constraints, enabling the model to acquire new skills while imitating the representations of the previous model trained for earlier tasks \cite{douillard2020podnet, kang2022class}. PodNet successfully implements restrictions by preserving the similarity of characteristics in the embedding space to tackle the problem of forgetting in continual learning (Douillard et al., 2020). In addition, iCaRL and End-to-End Incremental Learning techniques aid in preserving the model's memory of previous tasks by including the prediction probabilities of those tasks as soft goals during the training of new tasks \cite{rebuffi2017icarl, castro2018end}. 
However, some constraints may have empirical generalization issues and might not always transfer well across different tasks or domains, limiting their effectiveness in broader applications.

Actually, combining multiple of the above methods can further enhance the model's performance and robustness in handling incremental learning tasks. 

\section{Supplemenary Preliminary}
In this section, we first describe the problem setup of the CIL scenario. Then we present a naive experiment according to this setup, leading to the components of our method.

\textbf{Class-Incremental Learning Setup.} \ \ 
Unlike the traditional case where the model is exposed with random-shuffled training data of all classes, in the CIL scenario (considering the image-classification problem), the model observes a stream of class groups $\{\mathcal{Y}_s\}$ and corresponding training data $\{\mathcal{D}_s\}$ at each incremental step $s$. The incoming dataset $\{\mathcal{D}_s\}$ is in the form of ($x^s_i, y^s_i$) where $x^s_i$ is the input image and $y^s_i \in \mathcal{Y}_s$ is the label within label set $\mathcal{Y}_s$ of the $s$-th step. Additionally, the label space of all seen categories is denoted as $\hat{\mathcal{Y}}_s = \cup^s_{i=0}\mathcal{Y}_s$, where $\mathcal{Y}_s \cap \mathcal{Y}_{s'} = \emptyset$ for $s \neq s'$ typically.

Our method adopts the rehearsal strategy, which saves a portion (usually amounting to a certain number) of the trained data $\mathcal{M}_s $, a subset of $\cup^{s-1}_{i=0} \mathcal{D}_i$. At each incremental step $s$, the model is trained on $\hat{\mathcal{D}}_s = \mathcal{D}_s \cup \mathcal{M}_s$ and is required to perform well on all previously seen categories.

\textbf{DER.} \ \ 
DER \cite{yan2021dynamically}, the backbone framework we adopt, uses a dynamically expanding network to strike a balance between stability and plasticity. It has two stages in the learning process. In the representation learning stage, when a new task is being learned, the method freezes the previous group of feature extractors $\Phi_{s-1}$ and expands a novel extractor $\mathcal{F}_s$ to learn new features, where an auxiliary loss is used to promote learning.

\begin{equation}
    \label{der_feature}
    \Phi_{s}(x) = [\Phi_{s-1}(x), \mathcal{F}_{s}(x)]
\end{equation}

In the classifier learning stage, the method utilizes the current dataset to retrain the classifier. We apply this framework to SNNs but introduce additional components to further enhance its capability.

\textbf{SNNs Model.} \ \
Unlike ANNs, SNNs utilize spiking neurons to produce discrete 0/1 outputs. In this paper, we adopt the Leaky Integrate-and-Fire (LIF) model \cite{abbott1999lapicque}. The detailed dynamics of LIF neurons can be described as follows:
\begin{equation}
    \label{membrane_calculation}
    u^{t+1, n} = \tau u^{t, n} + \sum^{l(n-1)}_{j=1}{w^n o^{t+1, n-1}}
\end{equation}
\begin{equation}
    \label{output_calculation}
    o^{t+1, n} = \Theta(u^{t+1, n} - V_{th})
\end{equation}
\begin{equation}
    \label{membrane_resetting}
    u^{t+1, n} = u^{t+1, n} \cdot (1 - o^{t+1, n})
\end{equation}

where $\tau$ is the constant leaky factor, $u^{t, n}$ is the membrane potential at time $t$ in layer $n$, and $\sum^{l(n-1)}_{j=1}{w^n o^{t+1, n-1}}$ denotes the product of the synaptic weight and the spiking output of the previous layer. Neurons reset $u$ to 0 and emit a spike ($o = 1$) to the next layer when $u$ exceeds the firing threshold $V_{th}$. Here, $\Theta$ denotes the Heaviside step function. To enhance overall performance, we configure the last classifier to output the pre-synaptic inputs \cite{rathi2020diet, fang2021deep} without decay or firing.

To handle the non-differentiability of $\Theta$, \cite{wu2018spatio} introduces the method of surrogate gradient, and we apply a triangle-shaped function \cite{esser2016cover, rathi2020diet} as referenced in \cite{deng2021temporal}:
\begin{equation}
    \label{surrogate_function}
    \frac{\delta o^t}{\delta u^t} = \frac{1}{\gamma^2} max(0, \gamma - |u^t-V_{th}|)
\end{equation}

\section{Supplemenary Experimental settings}

\textbf{Benchmark Protocols} \ \ We test our method on two protocols for the CIFAR100 \cite{krizhevsky2009learning} benchmark. The first is \textit{CIFAR100-B0} \cite{rebuffi2017icarl}, which trains all 100 classes in specific splits over 5, 10 and 20 steps with a fixed memory size of 2000. The other is \textit{CIFAR100-B50} \cite{hou2019learning}, in which the model starts with training on 50 classes and subsequently learns the remaining classes in specific splits over 5 and 10 steps, using a fixed memory of 20 examples per class. We record the average incremental accuracy after each step (denoted as "Avg" as defined in \cite{rebuffi2017icarl}) and the final accuracy after the last step (denoted as "Last"). Each result in the table is the average of three orders (referring \cite{douillard2020podnet}) of the split dataset.

\textbf{Implementation Details} \ \ Our SNN model is implemented using spikingjelly \cite{fang2023spikingjelly} within the PyTorch framework. When training on CIFAR100, we use ResNet-19 \cite{fang2021deep} as the SNN feature extractor $\mathcal{F}_{s}$ at each step, which is a common setting for SNNs. Due to the scarcity of directly comparable SNN-related work, we primarily compare the performance of our method with that of several popular ANN methods using 18-layer ResNet or ViT as the basic network and one recently proposed SNN method. The results are directly taken from their respective papers. Furthermore, we adopt a herding strategy \cite{welling2009herding} to select the exemplars as memory at each step. For our SNN, we set the size of the time window to 4 when trained with CIFAR100.

\end{document}